\RequirePackage{fix-cm}
\documentclass[twocolumn, final]{svjour3}          
\smartqed  
\usepackage{graphicx}
\usepackage{amsmath}
\usepackage[noend]{algpseudocode}
\usepackage{array}
\usepackage{multirow}
\usepackage{subcaption}
\usepackage[linesnumbered,ruled]{algorithm2e}
\usepackage{xcolor}
\usepackage{soul}
\usepackage{fancyhdr}
\usepackage[pass]{geometry}
\usepackage{balance}
\usepackage{threeparttable}

\usepackage{gensymb}
\usepackage{color}
\usepackage{textcomp}
\newcommand{\mb}[1]{\textcolor{black}{#1}}

\begin{document}

\title{Run-time Mapping of Spiking Neural Networks to Neuromorphic Hardware}


\author{Adarsha Balaji \and Thibaut Marty \and Anup Das \and Francky Catthoor}


\institute{Adarsha Balaji \at Drexel University, Philadelphia, Pennsylvania, USA, 19104
              \email{adarsha.balaji@drexel.edu}           
           \and
           Thibaut Marty \at
              \email{thibaut.marty@ens-rennes.fr}
           \and
           Anup Das \at Drexel University, Philadelphia, Pennsylvania, USA, 19104
            \email{anup.das@drexel.edu}
            \and
           Francky Catthoor \at Neuromorphic Division, IMEC, 3001 Leuven,Belgium.
              \email{francky.catthoor@imec.be}
}


\maketitle

\begin{abstract}
Neuromorphic architectures implement biological neurons and synapses to execute machine learning algorithms with spiking neurons and \mb{bio}-inspired learning algorithms. These architectures are energy efficient and therefore, suitable for cognitive information processing on resource and power-constrained environments, ones where sensor and edge nodes of internet-of-things (IoT) operate. To map a spiking neural network (SNN) to a neuromorphic architecture, prior works have proposed design-time based solutions, where the SNN is first analyzed offline using representative data and then mapped to the hardware to optimize some objective functions such as minimizing spike communication \mb{or maximizing resource utilization.} \mb{In many emerging applications, machine learning models may change based on the input using some online learning rules. In online learning, new connections may form or existing connections may disappear at run-time based on input excitation.} Therefore, \mb{an already mapped} SNN may need to be re-mapped to the neuromorphic hardware to ensure optimal performance. Unfortunately, due to the high \mb{computation} time, design-time based approaches are not suitable for remapping \mb{a machine learning model} at run-time after every learning epoch.

In this paper, we propose a design methodology to partition and map the \mb{neurons} and synapses of online learning SNN-based applications to neuromorphic architectures at {run-time}. Our design methodology operates in two steps -- step 1 is a layer-wise greedy approach to partition SNNs into clusters of neurons and synapses incorporating the constraints of the neuromorphic architecture, and step 2 is a hill-climbing optimization algorithm that minimizes the total spikes communicated between clusters, improving energy consumption on the shared interconnect of the architecture. We conduct experiments to evaluate the feasibility of our algorithm using synthetic and realistic SNN-based applications. We demonstrate that our algorithm reduces SNN mapping time by an average 780x compared to a state-of-the-art design-time based SNN partitioning approach with only 6.25\% lower \mb{solution quality}.
\end{abstract}

\section{Introduction}\label{sec:Introduction}
Internet of things (IoT) is an emerging computing para\-digm that enables the integration of ubiquitous sensors over a wireless network \cite{fuqaha2015iot}. Recent estimates predict that over 50 billion IoT devices will be interconnected via the \mb{cloud} over the next decade \cite{evans2011internet}. In a conventional IoT, data collected from sensors and actuators are transferred to the cloud and processed centrally \cite{mohammadi2018deep}. However, with an increase in the number of connected IoT devices, processing on the \mb{cloud} becomes the performance and energy bottleneck \cite{shi2016promise}.    

Edge computing is emerging as a scalable solution to process large volumes of data by executing machine learning tasks closer to the data source e.g.  on a sensor or an edge node \cite{shi2016edge}. Processing on edge devices allows real-time data processing and decision making, and offers network scalability and privacy benefits as data transferred to the cloud over \mb{a possibly} insecure communication channel is minimized \cite{iot2017,mao2017mobile}. 

\mb{Spiking neural networks (SNNs) \cite{maass1997networks} are extremely energy efficient in executing machine learning tasks on event-driven neuromorphic architectures such as True\-North \cite{akopyan2015truenorth}, DYNAP-SE \cite{Moradi_etal18}, and Loihi \cite{davies2018loihi}, making them suitable for machine learning-based edge computing. A neuromorphic architecture is typically designed using $crossbars$, which can accommodate only a limited number of synapses per neuron to reduce energy consumption. To build a large neuromorphic chip, multiple crossbars are integrated using a shared interconnect such as network-on-chips \mb{(NoC)} \cite{benini2002networks}. To map an SNN to these architectures, the common practice is to partition the neurons and synapses of the SNN into clusters and map these clusters to the crossbars, optimizing hardware performance such as minimizing the number of spikes communicated between crossbar, which reduces energy consumption \cite{das2018mapping}.
}

Most prior works on machine learning-based edge computing focus on supervised approaches, where neural network models are first trained offline with representative data from the field and then deployed on edge devices to perform inference in real-time \cite{Shafique2017IOTML}. \mb{However,} data collected by IoT sensors constantly evolve \mb{over} time and may not resemble the representative data used to train the neural network model. This change in the relation between the input data and \mb{an} offline trained model is referred to as \textit{concept drift} \cite{conceptDrift2014Jo}. Eventually, the concept drift will reduce the prediction accuracy of the model over time, lowering its quality. Therefore, there is a clear need to periodically re-train the model using recent data with adaptive learning algorithms. \mb{Examples of such algorithms include} transfer learning \cite{pan2009survey}, lifelong learning \cite{thrun1998lifelong} and deep reinforcement learning\cite{mnih2015human}. 

Mapping decisions for a supervised SNN are made at design-time before the initial deployment of the trained model. However, in the case of online learning, when the model is re-trained, (1) synaptic connections within the \mb{SNN} may change, i.e. new connections may form and existing connection may be removed as new events are learned, and (2) weights of existing synaptic connections \mb{may} undergo changes after every learning epoch. In order to ensure the optimal hardware performance at all times, a \textit{run-time} approach is required that remaps the SNN to the hardware after every \mb{learning epoch}. Prior methods to partition and map an SNN to neuromorphic hardware, such as PSOPART\cite{das2018mapping}, SpiNeMap\cite{Balaji2019MappingHardwareb}, PyCARL\cite{Balaji2020PyCARLAP}, NEUTRAM\-S\cite{ji2016neutrams} and DFSynthesizer\cite{song2020compiling} are design-time approaches that require significant exploration time to generate a good solution. Although suitable for mapping supervised machine learning models, these approaches cannot be used at run-time to remap SNNs frequently.
For online learning, we propose an approach to perform run-time layer-wise mapping of SNNs on to crossbar-based neuromorphic hardware. \mb{The approach is implemen\-ted in two steps}. First, we perform a layer-wise greedy clustering of the neurons in the SNN. Second, we use an instance of hill-climbing optimization (HCO) to lower the total number of spikes communicated between the crossbars.

\textbf{Contributions}: Following are our key contributions. 
 \begin{itemize}
 	\item We propose an algorithm to partition and map online learning SNNs on to neuromorphic hardware for IoT applications in run-time;
 	\item We demonstrate suitability of our approach for online mapping in terms of the exploration time and total number of spikes communicated between the crossbars, when compared to a state-of-the-art design time approach.
 \end{itemize}

 The remainder of this paper is organized as follows, Section \ref{sec:RelatedWorks} presents the background, Section \ref{sec:Methodology} discusses the problem of partitioning a neural network into clusters to map on to the crossbars neuromorphic hardware and describes our two-step approach. Section \ref{sec:Results} presents the experimental results based on synthetic applications. Section \ref{sec:Conclusion} concludes the paper followed by a discussion in Section \ref{sec:Discussion}.
 
 \begin{figure*}[t!]
	\centering
	\centerline{\includegraphics[width=0.99\textwidth]{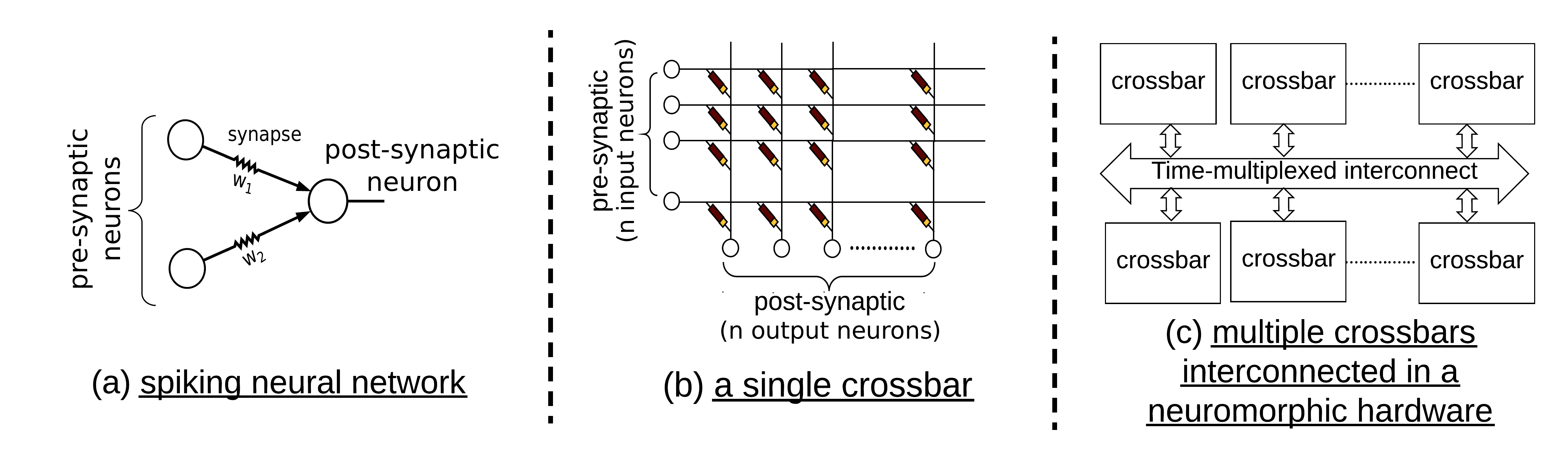}}
	\caption{Overview of a SNN hardware: (a) connection of pre- and post-synaptic neurons via synapses in a spiking neural network, (b) a crossbar organization with fully connected pre- and post-synaptic neurons, and (c) a modern neuromorphic hardware with multiple crossbars and a time-multiplexed interconnect.}
	\label{fig:crossbar}
\end{figure*}

\section{Background}\label{sec:RelatedWorks}

Spiking neural networks are event-driven computational models inspired by the mammalian brain. Spiking neurons are typically implemented using Integrate-and-Fire (I\&F) models \cite{chicca2003vlsi} and communicate using short impulses, called \emph{spikes}, via synapses. \mb{Figure \ref{fig:crossbar}(a) illustrates an SNN with \textit{two} pre-synaptic neurons connected to a post-synaptic neuron via synaptic elements with weights $w1$, $w2$ respectively. When a pre-synaptic neuron generates a spike, current is injected into the post-syna\-ptic neuron, proportional to the product of the spike voltage and the conductance of the respective synapse.} SNNs are trained by adjusting the synaptic weights using a supervised, a semi-supervised, or an unsupervised approach \cite{kasabov2001evolving,lee2016training,mostafa2018supervised}. 



Due to the ultra-low power footprint of neuromorphic hardware, several machine learning applications based on SNNs are implemented. In \cite{das2018heartbeat}, the authors propose a multi-layer perceptron (MLP) based SNN to classify heartbeats using electrocardiagram (ECG) data. In \cite{diehl2015unsupervised}, the authors propose the handwritten digit recognition
using unsupervised SNNs. In \cite{das2017unsupervised}, a spiking liquid state machine for heart-rate estimation is proposed. A SNN-based liquid state machine (LSM) for facial recognition is proposed in \cite{grzyb2009facial}. In \cite{Balaji2018Power-AccuracyHardware}, the authors propose a technique to convert a convolutional neural network (CNN) model for heartbeat classification into a SNN, with a minimal loss in accuracy.

Typically, SNNs are executed on special purpose neuromorphic hardware. These hardware can (1) reduce energy consumption, due to their low-power designs, and (2) improve application throughput, due to their distributed computing architecture. Several digital and mixed-signal neuromorphic hardware are \mb{recently developed} to execute SNNs, such as Neurogrid\cite{neurogrid2014}, TrueNorth \cite{Akopyan2015TrueNorth:Chip} and DYNAP-SE \cite{Moradi2018ADYNAPs}. Although these hardware differ in their operation (analog vs. digital), they all support crossbar-based architectures. A crossbar is a two-dimensional arrangement of synapses ($n^2$ synapses for $n$ neurons). \-\mb{Figure \ref{fig:crossbar}(b) illustrates a single crossbar with $n$ pre-synaptic neurons and $n$ post-synaptic neurons. The pre- and post-synaptic neurons are connected via synaptic elements.}  Crossbar size (n) is limited ($<$512) as scaling the size of the crossbar will lead to an exponential increase in dynamic and leakage energy. Therefore, to build large neuromorphic hardware, multiple crossbars are integrated using a shared interconnect, \mb{as illustrated in Figure \ref{fig:crossbar}(c)}.



In order to execute an SNN on a neuromorphic hardware, the SNN is first partitioned into clusters of neurons and synapses. The clustered (local) synapses are then mapped to the crossbars and the inter-cluster syna\-pses to the time-multiplexed interconnect. Several design time partitioning approach are presented in literature. \mb{In \cite{xia2019memristive,wijesinghe2018all,wen2015eda} the authors proposes techniques to efficiently map the neurons and synapses on a crossbar. The aim of these techniques is to maximize the utilization of the crossbar. NEUTRAMS partitions the SNN for crossbar-based neuromorphic hardware \cite{Ji2016NEUTRAMS:Constraints}. The NEUTRAMs approach also looks to minimize the energy consumption of the neuromorphic hardware executing the SNN. PyCARL \cite{Balaji2020PyCARLAP} facilitates the hardware-software co-simulation of SNN-based applications. The framework allows users to analyze and optimize the partitioning and mapping of an SNN on cycle-accurate models of neuromorphic hardware. DFSynthesizer \cite{song2020compiling} uses a greedy technique to partition the neurons and synapses of an SNN. The SNN partitions are mapped to the neuromorphic hardware using an algorithm that adapts to the available resources of the hardware. SpiNe\-Map \cite{Balaji2019MappingHardwareb} uses a greedy partitioning technique to partition the SNN followed by a meta-heuristic-based technique to map the partitions on the hardware. PSO\-PART SNNs to a crossbar architecture \cite{mappingSNN2018das}. The objective of SpiNe\-Map and PSOPART is to minimize the spike communication on the time-multiplexed interconnect in order to improve the overall latency and power consumption of the DYNAP-SE hardware.} \mb{Table \ref{tab:contributions} compares our contributions to the state-of-the-art techn\-iques.}

\begin{table}[t!]
	\renewcommand{\arraystretch}{1}
	\setlength{\tabcolsep}{6pt}
	\centering
	{\fontsize{7}{10}\selectfont
		\begin{tabular}{m{2.5cm}|m{1.4cm}p{2.8cm}}
		\hline%
			\centering\textbf{Related Works} & \textbf{Run-time Mapping} & \textbf{Objective}\\
			\hline
			\hline
			\centering\cite{xia2019memristive,wijesinghe2018all,wen2015eda} & \centering $\times$ & Maximize single crossbar utilization\\
			 \centering NEUTRAMS \cite{ji2016neutrams} & \centering $\times$  & Minimize number of crossbars utilized \\
			 \centering SpiNeMap \cite{Balaji2019MappingHardwareb} & \centering $\times$  & Minimize spikes on time-multiplexed interconnect\\
			 \centering PSOPART \cite{das2018mapping} & \centering $\times$  & Minimize spikes on time-multiplexed interconnect\\
			\centering DFSynthesizer \cite{song2020compiling} & \centering $\times$  & Optimize the hardware utilization in run-time\\
			\hline
			\hline
			\centering\textcolor{blue}{Proposed} & \centering \textcolor{blue}{$\surd$} & \textcolor{blue}{Reduces energy consumption of online learning SNNs on hardware.}\\
			\hline
	\end{tabular}}
	\caption{Summary of related works.}
	\label{tab:contributions}
\end{table}
\begin{figure*}[t!]
	\centering
	\centerline{\includegraphics[width=0.99\textwidth]{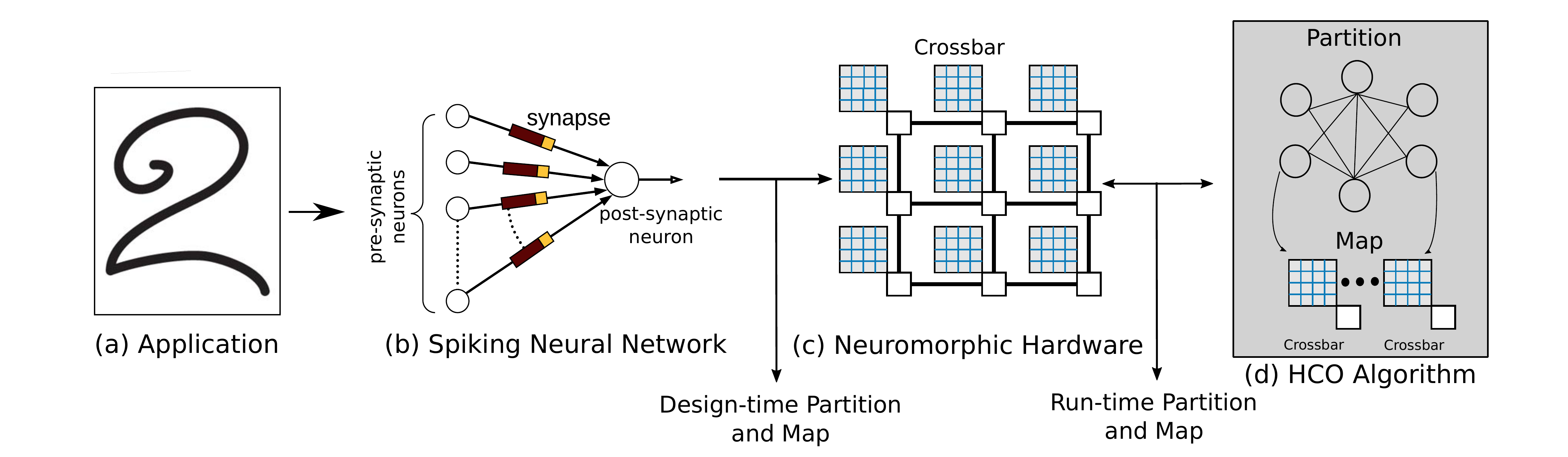}}
	\caption{Mapping of online learning SNN on Neuromorphic Hardware.}
	\label{fig:designFlow}
\end{figure*}

As these partitioning approaches aim to find the optimal hardware performance, their exploration time is relatively large and therefore not suitable for partitioning and re-mapping of online learning SNNs. \mb{Run-time approaches are proposed for task mapping on multiprocessor systems. A heuristic-based run-time manager is proposed in \cite{das2015runtime}. The run-time manager controls the thread allocation and voltage/frequency scaling for energy efficient execution of applications on multi processor systems. In \cite{mahmood2017energy}, the authors propose a genetic algorithm-based run-time manager to schedule real-time tasks on Dynamic Voltage Scaling (DVS) enabled processors, with an aim to minimize energy consumption. A workload aware thread scheduler is proposed in \cite{Dhiman2009PDRAM:System} for multi-processor systems. In \cite{das2015workload}, the authors propose a multinomial logistic regression model to partition the input workload in run-time. Each partition is then executed at pre-determined frequencies to ensure minimum energy consumption. In \cite{das2012fault}, the authors propose a technique to remap tasks run on faulty processors with a minimal migration overhead. A thermal-aware task scheduling approach is proposed in \cite{cui2012fast} to estimate and reduce the temperature of the multi processor system at run-time. The technique performs an extensive design-time analysis of fault scenarios and determines the optimal mapping of tasks in run-time. However, such run-time techniques to remap SNN on neuromorphic hardware are not proposed. To the best of our knowledge, this is the first work to propose a run-time mapping approach with a significantly lower execution time when compared to existing design-time approaches. Our technique reduces the spikes communicated on the time-multiplexed interconnect, therefore reducing the energy consumption.}

 \begin{figure}[h!]
	\centering
	\centerline{\includegraphics[width=0.99\columnwidth, height=2.5cm]{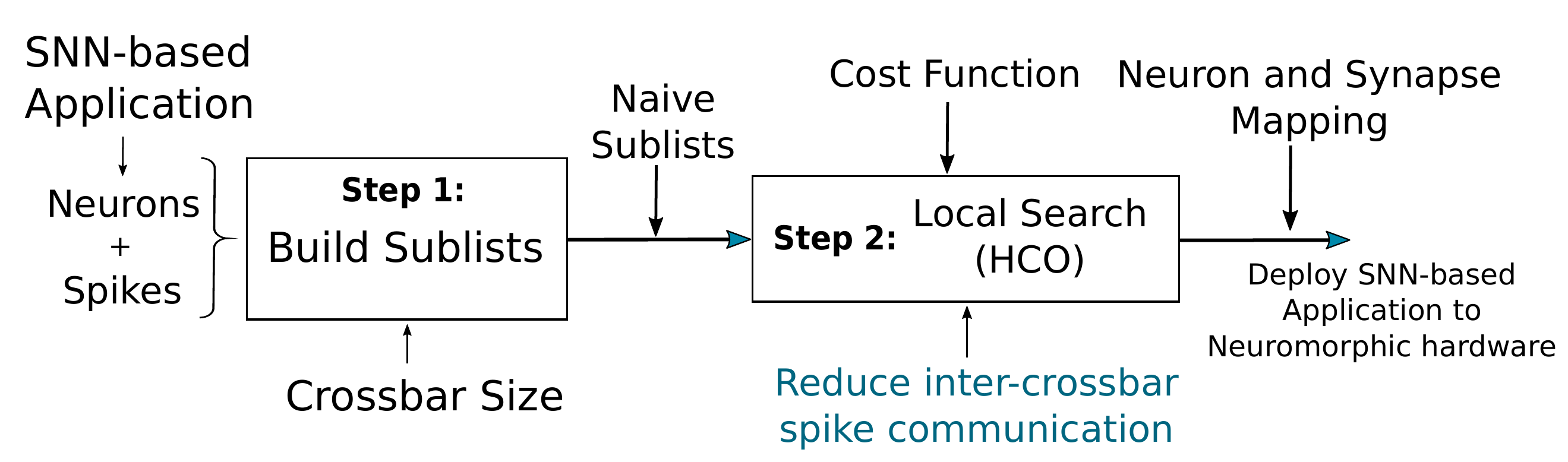}}
	\caption{Overview of proposed partitioning algorithm.}
	\label{fig:algo}
\end{figure}

\section{Methodology}\label{sec:Methodology}
\mb{The proposed method to partition and map an SNN in run-time is illustrated in Figure \ref{fig:designFlow} illustrates. The network model is built using a directed graph, wherein each edge represents a synapse whose weight is the total number of spikes communicated between the two SNN neurons.} The input to the mapping algorithm is a list of all the neurons \textit{(A)}, the total number of spikes communicated over each synapse and the size of a crossbar \textit{(k)}. The mapping algorithm is split into two steps, as shown in Figure \ref{fig:algo}. 

\begin{figure}[h!]
	\centering
	\centerline{\includegraphics[width=0.99\columnwidth]{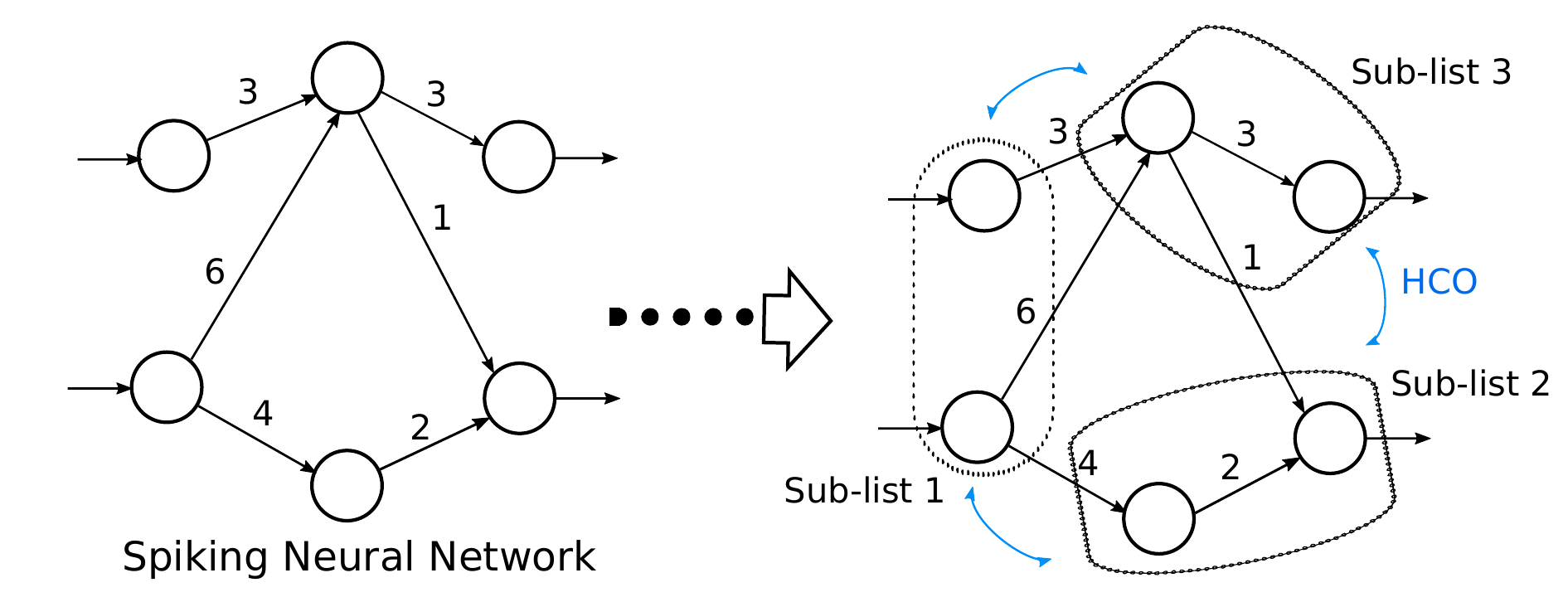}}
	\caption{Partitioning of an SNN.}
	\label{fig:clustering}
\end{figure}

Figure \ref{fig:clustering} illustrates the partitioning of an SNN with 6 neurons into 3 sub-lists. The spikes communicated between the neurons is indicated on the synapse. First, we divide the input list of neurons into sub-lists (Section \ref{subsec:sublists}), such that each sub-list can be mapped to an available crossbar. Second, we reduce the number of spikes communicated between the sub-lists (Section \ref{subsec:search}), by moving the neurons between the sub-list (indicated in blue).
 
\subsection{Building Sub-lists}\label{subsec:sublists}
Algorithm \ref{alg:part_one} describes the greedy partitioning approach. The objective is to greedily cut the input list of neurons \emph{(A)} into \textit{s} sub-lists, where \textit{s} is the total number of crossbars in the given design. The size of a sub-list is determined by the size of the crossbars \textit{(k)} on the target hardware. A variable \textit{margin} \textit{(line 3)} is defined to store the unused neuron slots available in each sub-list. The \textit{mean} \textit{(line 4)} number of spikes generated per crossbar is computed using the total number of spikes communicated in the SNN-based application. A \textit{cost} function (Algorithm \ref{alg:part_three}) is defined to compute the total number of spikes communicated (cost) between each of the sub-lists. 

The algorithm iterates over the neurons ($n_i$) in the input list ($A$) and updates the slots in the current sub-list \textit{(line 8)}. Neurons are added to the current sub-list until one of following two criteria are met - (1) the length of the sub-list equals \textit{k}, or (2) the cost (number of spikes) is greater than the \textit{mean} value and sufficient extra slots $(margin)$ are still available. When the criteria is met, the current sublist is validated and its boundary stored. When the penultimate sub-list is validated, the execution ends because the boundary of the last sub-lists is already known (nth element in list). The list p contains the sub-lists boundaries.

\begin{algorithm}[t!]
	\small
		\textbf{procedure} FUNCTION $(A[1 \to n])$ \\
	    \ForEach{Crossbar s $\in$ $p$}
	    {
	        \tcc{iterate over all crossbars in p}
		    \textbf{Input} the variable margin\;
	        \tcc{Mean spikes per crossbar}
		    \textbf{Compute} Mean\;
	        \tcc{iterate over all neurons in A}
		    \ForEach{$n_i$ $\in$ $A$}
		    {
		        \tcc{Cost is the number of spikes in current cluster}
		        \textbf{Compute} Cost\;
		        \While{Cost $\leq$ Mean}
		        {
		    	    \textbf{Assign} $n_i$ to crossbar p\;
		        }
	        }
	    }
	\caption{Building Sublists}
	\label{alg:part_one}
\end{algorithm}

\begin{algorithm}[t]
	\small
	{
	    \textbf{procedure} FUNCTION $(A[1 \to n], p[1 \to s])$\\
	    $max \leftarrow 0$\;
	    \ForEach{Cluster (p[i])}
	    {
	        $sum \leftarrow 0$ \;
	        \ForEach{n in p[i]}
	        {
	            \tcc{total spikes communicated}
	            \textbf{compute} Sum\;
	        }
	        \If {Sum $>$ Max}
	        {
	            Max $\leftarrow$ Sum\;
            }
	    }
	}
	\caption{Cost Function.}
	\label{alg:part_three}
\end{algorithm}

\begin{algorithm}[t!]
	\small{
		\textbf{procedure} FUNCTION $(A[1 \to n], p[1 \to s])$\\
	    \tcc{compute the initial cost}		
		\textbf{compute} Cost\;
		\ForEach{$n$ in A}{
			\textbf{move} $n$ across cluster boundary\;
			\textbf{compute} new Cost $C_n$\;
	    \textbf{select} min($C_n$)\;
		}
	\tcc{end 2-part procedure}
	}
	\caption{Hill Climbing Algorithm.}
	\label{alg:two_part_full}
\end{algorithm}

\subsection{Local Search}\label{subsec:search}
The solution obtained from Algorithm-1 is naive and not optimal. Although each sublist $s$ obtained from Algori\-thm-1 meets the cost criteria, it is possible to have unevenly distributed costs across the sublists. We search for a better solution by performing multiple local searches to balance the cost. This is done by using the hill-climbing optimization technique to iterate through the sublist and \textit{move} its boundary.

Algorithm-\ref{alg:two_part_full} describes the hill-climbing optimization technique. The technique relies on a \textit{cost function} \textit{(line 2)} to compute and evaluate a solution. The cost function used in the optimization process is shown in Algo\-rithm-\ref{alg:part_three}. The cost function computes the maximum cost (number of spikes) for a chosen sub-list. The optimal solution should contain the lowest cost. The algorithm iterates through each subslist to search for the best solution (cost) of its neighbors. The algorithm begins by moving the boundary of a sub-list one position to the left or one position to the right. Each neuron ($n_i$) in the sublist is moved across the boundary to a neighboring sub-list and the \textit{cost} of the neighbors are computed. The algorithm selects the solution with the local minimum cost. The process is repeated for every neuron in the list (A) until the sub-lists with the minimum cost is found.

\section{Evaluation}\label{sec:Evaluation}
\subsection{Simulation environment}
We conduct all experiments on a system with 8 CPUs, 32GB RAM, and NVIDIA Tesla GPU, running Ubuntu 16.04. 

\begin{itemize}
	\item \textbf{CARLsim} \cite{Chou2018CARLsim4} : A GPU accelerated simulator used to train and test SNN-based applications. CARLsim reports spike times for every synapse in the SNN. 
	\item  \textbf{DYNAP-SE} \cite{Moradi_etal18}: Our approach is evaluated using the  DYNAP-SE model, with 256-neuron crossbars interconnected using a NoC. \cite{zhao2006new}.
\end{itemize}

\subsection{Evaluated applications}
In order to evaluate the online mapping algorithm, we use 2 synthetic and 2 realistic SNN-based applications. Synthetic applications are indicated with an 'S\_' followed by the number of neurons in the application. Edge detection (EdgeDet) and MLP-based digit recognition (MLP-MNIST) are the two realistic applications used. Table \ref{tab:apps} also indicates the number of synapses (column 3), the topology (column 4) and the number of spikes for the application obtained through simulations using CARLsim \cite{Chou2018CARLsim4}.
\begin{figure*}[h!]
	\centering
	\centerline{\includegraphics[width=0.7\textwidth]{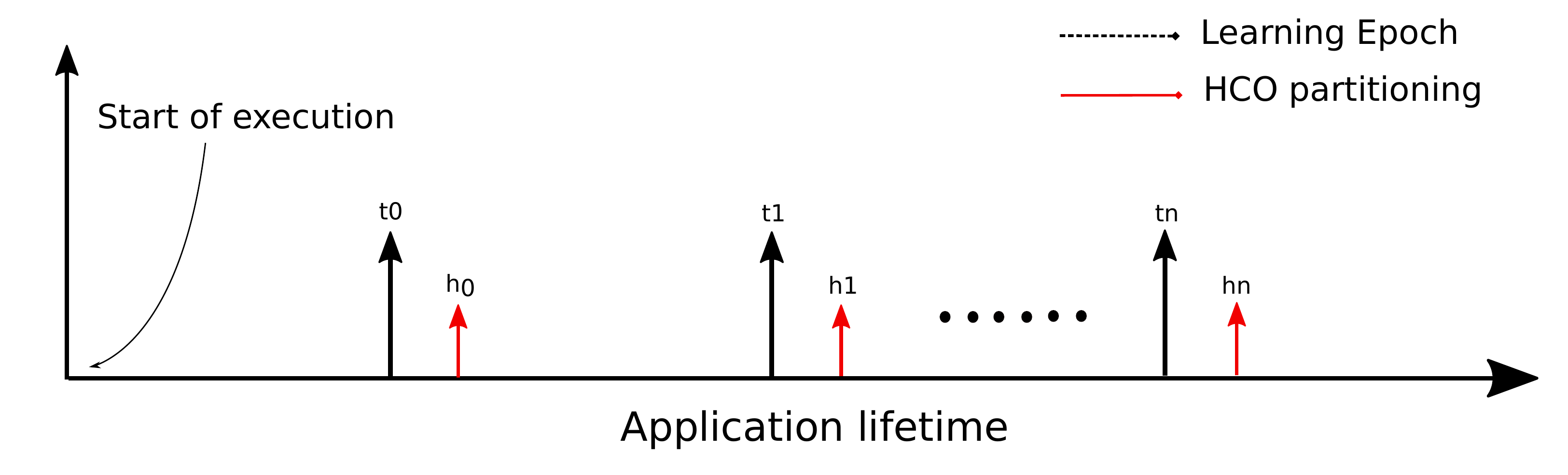}}
	\caption{Life-time of online learning SNN}
	\label{fig:timing}
\end{figure*}

\begin{figure*}[t!]
	\centering
	\centerline{\includegraphics[width=0.55  \textwidth]{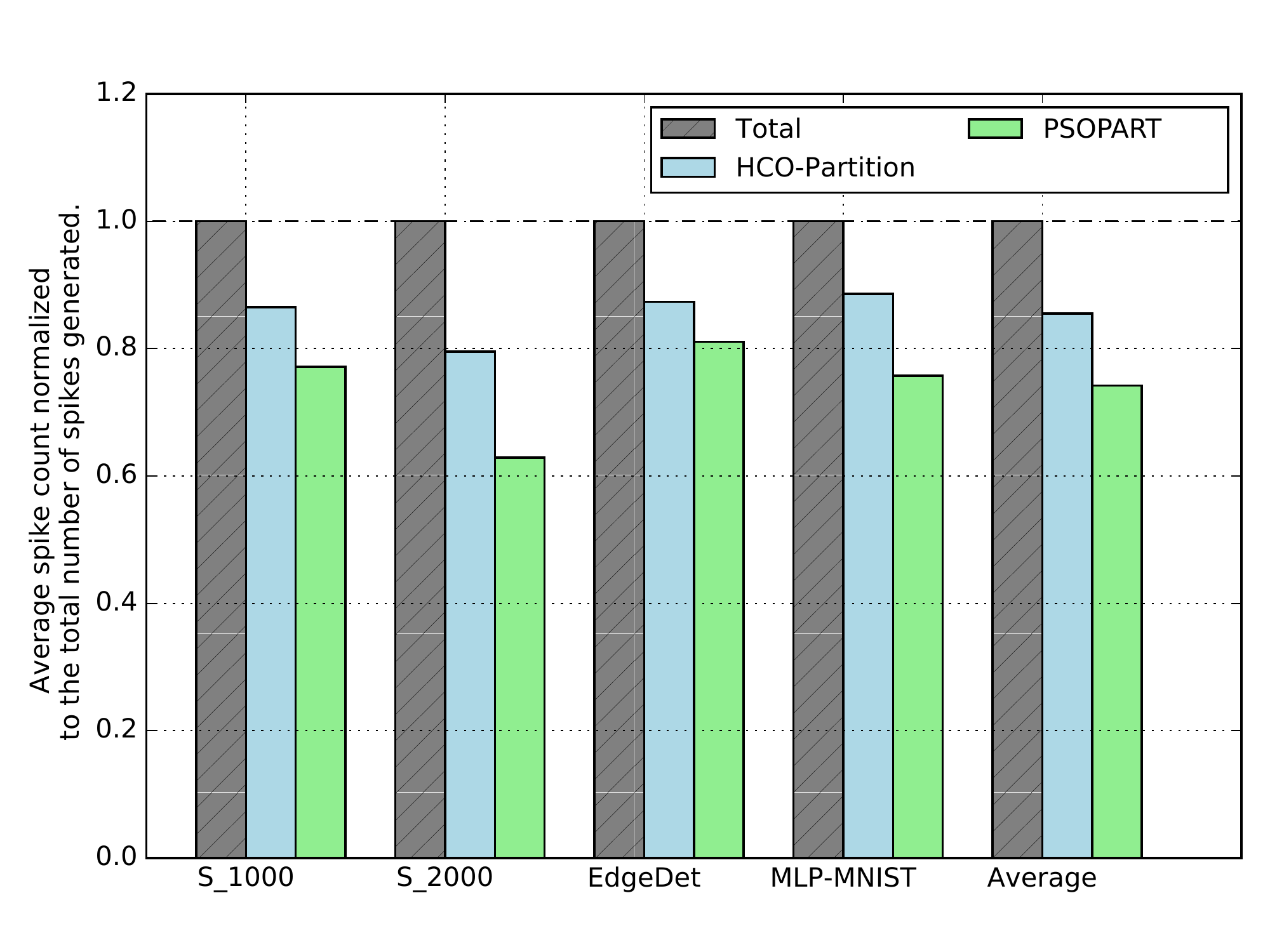}}
	\caption{Number of spikes communicated on the time-multiplexed interconnect normalized to the total number of spikes generated.}
	\label{fig:spikes}
\end{figure*}

\begin{table}[t!]
    \resizebox{\columnwidth}{!}{
	\renewcommand{\arraystretch}{1}
	\setlength{\tabcolsep}{2pt}
	\centering
	\fontsize{30}{60}\selectfont
		\begin{tabular}{cc|c|l|c}
			\hline
			\textbf{Category} & \textbf{Applications} & \textbf{ Synapses } & \textbf{Topology} & \textbf{Spikes}\\
			\hline
			\multirow{2}{*}{synthetic} & S\_1000 & 240,000 & FeedForward (400, 400, 100) & 5,948,200\\
			& S\_2000 & 640,000 & FeedForward (800, 400, 800) & 45,807,200\\
			\hline
			\multirow{2}{*}{realistic} &  EdgeDet \cite{Chou2018CARLsim4} & 272,628 & FeedForward (4096, 1024, 1024, 1024) & 22,780\\
			& MLP-MNIST \cite{diehl2015unsupervised} & 79,400 & FeedForward (784, 100, 10) & 2,395,300\\
			\hline
	\end{tabular}}
	\caption{Applications used for evaluating.}
	\label{tab:apps}
\end{table}

\subsection{Evaluated design-time vs run-time approach}
In order to compare the performance of our proposed run-time approach, we choose a state-of-the-art design-time approach as the baseline. The crossbar size for both the algorithms is set to 256 (k=256).In this paper we compare the following approaches: 

\begin{itemize}
    \item \emph{PSOPART} \cite{das2018mapping}: The PSOPART approach is a design-time  partitioning technique that uses and instance of particle swarm optimization (PSO) to minimize the number of spikes communicated on the time-multiplexed interconnect.   
    
    \item \emph{HCO-Partitioning}: Our HCO-partitioning approach is a two-step layer-wise partitioning technique with a greedy partitioning followed by a HCO-based local search approach to reduce the number of spikes communicated between the crossbars. 
    
\end{itemize}

\section{Results}\label{sec:Results}
Table \ref{runtime} reports the execution time (in seconds) of the design-time and run-time mapping algorithms for synthetic and realistic applications, respectively. 
We make the following two observations. \textit{First}, on average, our HCO partitioning algorithm has an execution time 780x lower than that of the PSOPART algorithm. Second, the significantly lower run-time of the HCO partitioning algorithm ($<$50 seconds) allows for the online learning SNN to be re-mapped on the edge devices, before the start of the next training epoch. 

\begin{table}[h]
    \resizebox{\columnwidth}{!}{
	\renewcommand{\arraystretch}{1.1}
	\setlength{\tabcolsep}{2pt}
	\centering
	{\fontsize{8}{12}\selectfont
		\begin{tabular}{cc|c|c}
			\hline
			\textbf{Category} & \textbf{Applications} & \textbf{PSOPART (sec)} & \textbf{HCO-Partition (sec)}  \\
			\hline
			\multirow{2}{*}{synthetic} & S\_1000 & 20011.33
 & 19.10\\
			& S\_2000 & 45265.00 & 24.68 \\
			\hline
			\multirow{2}{*}{realistic} & EdgeDet & 6771.02 & 45.62 \\
			& MLP-MNIST & 5153.41 & 11.03 \\
			\hline
	\end{tabular}}}
	\caption{Execution time of design-time and proposed run-time approach in seconds.}
	\label{runtime}
\end{table}

Figure \ref{fig:timing} shows the lifetime of an online learning application with respect to the execution times of each training epoch (t) and the HCO partitioning algorithm (h). The execution time of the partitioning algorithm needs to be significantly lower than the time interval between training epochs. This is achieved with the HCO-partitioning algorithm as its execution time is significantly (780x) lower than the state-of-the-art design-time approaches. 

In Figure \ref{fig:spikes}, we compare the number of spikes communicated between the crossbars while partitioning the SNN using the HCO partitioning algorithm when compared to the design-time PSOPART approach. We see that, on average, the PSOPART algorithm reduces the number of spikes by a further ~6.25\%, when compared to the HCO partitioning algorithm. The PSOPART will contribute to a further reduction in the overall energy consumed on the neuromorphic hardware. However, this outcome is expected as the design-time partitioning approach is afforded far more exploration time to minimize the number of spikes communicated between the crossbars. Also, the effects of \textit{concept drift} will soon lead to the design-time solution becoming outmoded. Therefore, a run-time partitioning and re-mapping of the SNN will significantly improve the performance of the SNN on the neuromorhpic hardware and mitigate the effects of \textit{concept drift}.

\section{Conclusion}\label{sec:Conclusion}
In this paper, we propose an algorithm to re-map online learning SNNs on neuromorphic hardware. Our approach performs the run-time mapping in two steps: (1) a layer-wise greedy partitioning of SNN neurons, and (2) a hill-climbing based optimization of the greedy partitions with an aim to reduce the number of spikes communicated between the crossbars. We demonstrate the in-feasibility of using a state-of-the-art design-time approach to re-map online learning SNNs in run-time. We evaluate the our approach using synthetic and realistic SNN applications. Our algorithm reduces SNN mapping time by an average 780x when compared to a state-of-the-art design-time approach with only 6.25\% lower performance.

\section{Discussion}\label{sec:Discussion} 
In this section we discuss the scalability of our approach. Each iteration of Algorithm-1 performs basic math operations. The hill-climbing algorithm computes as many as 2x(s-2) solutions, and performs a comparison to find the minimum cost across all the solutions. In our case, the co-domain of the cost function are well-ordered positive integers. The cost function is also linear in $n$, however the hill-climb optimization algorithm only terminates when the local minimum cost function is computed. Therefore, it is in our interest to optimize the number of times the cost function is to be run.

\section*{Acknowledgment}
\mb{This work is supported by 1) the National Science Foundation Award CCF-1937419 (RTML: Small: Design of System Software to Facilitate Real-Time Neuromorphic Computing) and 2) the National Science Foundation Faculty Early Career Development Award CCF-1942697 (CAREER: Facilitating Dependable Neuromorphic Computing: Vision, Architecture, and Impact on Programmability).}


%
%

\bibliographystyle{spmpsci}
\bibliography{snnhw,sch_anup,mendeley}

\end{document}